\documentclass[letterpaper, 10 pt, conference]{IEEEtran}
\IEEEoverridecommandlockouts 

\usepackage{cite}
\usepackage{graphics} 
\usepackage{epsfig} 
\usepackage{mathptmx} 
\usepackage{times} 
\usepackage{amssymb}  
\usepackage[cmex10]{amsmath}
\usepackage{amsthm}
\usepackage{amsfonts}
\usepackage{latexsym}
\usepackage{mathrsfs}
\usepackage{xcolor}
\usepackage[normalem]{ulem}
\usepackage{stfloats}
\usepackage{mathtools}
\usepackage{wrapfig}
\usepackage{xspace}
\usepackage{pbox}
\usepackage{geometry}
\usepackage{graphicx}
\usepackage{placeins}
\usepackage{flushend}

\ifCLASSOPTIONcompsoc
    \usepackage[caption=false, font=normalsize, labelfont=sf, textfont=sf]{subfig}
\else
\usepackage[caption=false, font=footnotesize]{subfig}
\fi

\definecolor{wdc}{RGB}{34,139,34}

\definecolor{cal}{RGB}{255,110,30}

\title{\LARGE \bf
Stabilization of Exoskeletons through Active Ankle Compensation
}

\author{Thomas Gurriet$^{1}$, Maegan Tucker$^{1}$, Claudia Kann$^{1}$, Guilhem Boeris$^{2}$, and Aaron D. Ames$^{1}$
\thanks{*This work has been conducted under IRB No. 16-0693.}%
\thanks{$^{1}$Dept. of Mechanical and Civil Engineering, California Institute of Technology, Pasadena, CA 91125.}%
\thanks{$^{2}$Wandercraft SAS, Paris, France.}
}

\begin{document}

\maketitle
\thispagestyle{empty}
\pagestyle{empty}

\begin{abstract}
This paper presents an active stabilization method for a fully actuated lower-limb exoskeleton. The method was tested on the exoskeleton ATALANTE, which was designed and built by the French start-up company Wandercraft. The main objective of this paper is to present a practical method of realizing more robust walking on hardware through active ankle compensation. The nominal gait was generated through the hybrid zero dynamic framework. The ankles are individually controlled to establish three main directives; (1) keeping the non-stance foot parallel to the ground, (2) maintaining rigid contact between the stance foot and the ground, and (3) closing the loop on pelvis orientation to achieve better tracking. Each individual component of this method was demonstrated separately to show each component's contribution to stability. The results showed that the ankle controller was able to experimentally maintain static balance in the sagittal plane while the exoskeleton was balanced on one leg, even when disturbed. The entire ankle controller was then also demonstrated on crutch-less dynamic walking. During testing, an anatomically correct manikin was placed in the exoskeleton, in lieu of a paraplegic patient. The pitch of the pelvis of the exoskeleton-manikin system was shown to track the gait trajectory better when ankle compensation was used. Overall, active ankle compensation was demonstrated experimentally to improve balance in the sagittal plane of the exoskeleton-manikin system and points to an improved practical approach for stable walking.

\end{abstract}              
\section{Introduction}
\label{sec:Introduction}
Active lower-limb exoskeleton technology has the potential to benefit approximately 6.7 million people in the United States who are limited by the effects of stroke, polio, multiple sclerosis, spinal cord injury, and cerebral palsy \cite{HerrActiveOrthoses}. The term ``exoskeleton'' is used to describe a mobility or rehabilitation device that augments the power of existing joints. While the term is traditionally associated with devices that assist physically challenged persons \cite{chen2015design}, \cite{esquenazi2012rewalk}, \cite{neuhaus2011design},
\cite{swift2011control}, exoskeleton can also be designed to improve strength and endurance of able-bodied persons \cite{krut2010moonwalker}, \cite{zoss2006biomechanical}. For lower-limb exoskeletons, this augmentation occurs on the majority of the joints of the lower extremities. An active device directly controls the joints rather than depending on simple mechanical coupling.  According to a survey taken by wheelchair users and healthcare professionals, the main benefit of exoskeleton technology is the resulting health benefits \cite{Wolff2014}. These health benefits primarily include pressure relief, increased circulation, improved bone density, improved bowel and bladder function, and reduced risk of orthostatic hypotension. The most important exoskeleton design consideration indicated by survey takers was ``minimizes the risk of falling'', a strong indication that fall prevention and robustness of stability should be the focus of exoskeleton research.
\begin{figure}[t!]
    \centering
   \subfloat[][Manikin Exoskeleton System]{\includegraphics[width=.5\columnwidth]{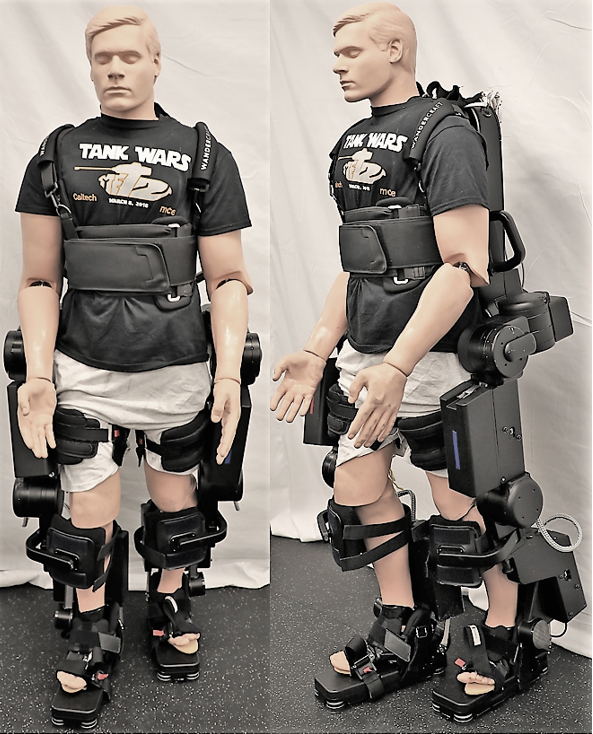}}
   \subfloat[][Kinematic Model]{\includegraphics[width=.335\columnwidth]{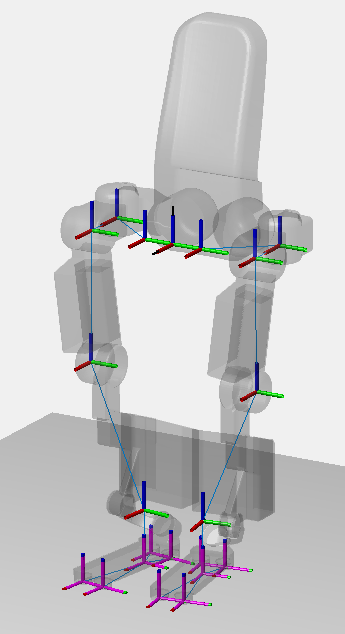}}\\
    \caption{Manikin Exoskeleton System}
    \label{fig:randy}
\end{figure}\\
\indent While research in this area began in the 1960s, until recently most lower-limb exoskeletons relied on the use of crutches to maintain balance and change direction \cite{HerrExoSurvey}. In addition, gaits were frequently slow and considered static. There was a notable shift when Wandercraft first introduced their exoskeleton ATALANTE  \cite{feedbackControlofanExoskeleton}, shown in Fig. \ref{fig:randy}. ATALANTE is unique in that it was the first demonstration of crutch-less walking for paraplegics through utilizing methods of partial hybrid zero dynamics to formally generate gaits tailored for each user \cite{restoringLocomotion}. The hybrid zero dynamics method has been implemented on many robotic platforms to achieve dynamic walking that is provably stable as well as experimentally realizable \cite{RESCLF}, \cite{GrizTerrain}, \cite{AyongaHZD}, \cite{reher2016realizing}. Even though the generated trajectories are provably stable they are susceptible to practical issues such as hardware flexibility and uneven terrain. This paper demonstrates a practical method of achieving even more robust dynamic walking through direct control of the ankles. The objective of the ankle controller is broken down into three components: (1) keeping the non-stance foot parallel to the ground, (2) maintaining rigid contact between the stance foot and the ground, and (3) closing the loop on pelvis orientation to achieve better tracking. Each individual component of this method was demonstrated separately to show each components contribution to stability.\\
\indent The remainder of the paper is organized as follows. In Sec. \ref{sec:GaitGeneration} the initial gait generation, in line with previous work, is briefly discussed. Following that, in Sec. \ref{sec:ProposedMethod}, the drawbacks of the existing methods are explored and a proposed solution is outlined. In Section \ref{sec:stableTrajectory} the proposed ankle controller is applied to dynamic walking and the results are presented.          
\section{Gait Generation}
\label{sec:GaitGeneration}

\subsection{Mechanical Model}

ATALANTE is a lower-limb exoskeleton designed by the French startup company Wandercraft. It is intended to be used by paraplegics in medical center settings for rehabilitation. For the purpose of this study an anatomically correct 145 lb dummy was used instead of a patient, seen in Figure \ref{fig:randy}. This was in order to ensure the framework worked without the stabilizing aid of a human patient, who frequently have the ability to rescue a gait with their torso.  

The exoskeleton has 12 degrees of freedom - 3 hip joints, 1 knee joint, and 2 ankle joints per leg - plus there are an additional 6 floating base coordinates located at the pelvis link. Assuming the pelvis link has Cartesian position $p_b \in \mathbb{R}^3$ and orientation $\phi_b \in SO(3)$ with respect to the world frame and $q_i \in \mathcal{Q}_i \subset \mathbb{R}^{12}$ represent the 12 actuated joints then the total space of the system can be represented as:
\begin{equation}
    q = \{p_b, \phi_b, q_i \} \in \mathbb{R}^3 \times SO(3) \times \mathcal{Q}_i = \mathcal{Q}
\end{equation}
While the ankle is driven by a special mechanism, all other degrees of freedom are independently actuated by brushless DC motors. A digital encoder is mounted on each motor to estimate joint position and velocity. 3-axis load cells in the feet allow for impact detection and center of pressure calculations. The exoskeleton is controlled by a central computer board running a real-time operating system and in charge of all high-level computations. The manikin is secured to the exoskeleton by fasteners at the ankle, shin, thigh, and abdomen. ATALANTE is self-powered with a battery pack.

\subsection{Gait Generation}
\begin{figure}[t]
    \centering
    \includegraphics[width = 0.85\columnwidth]{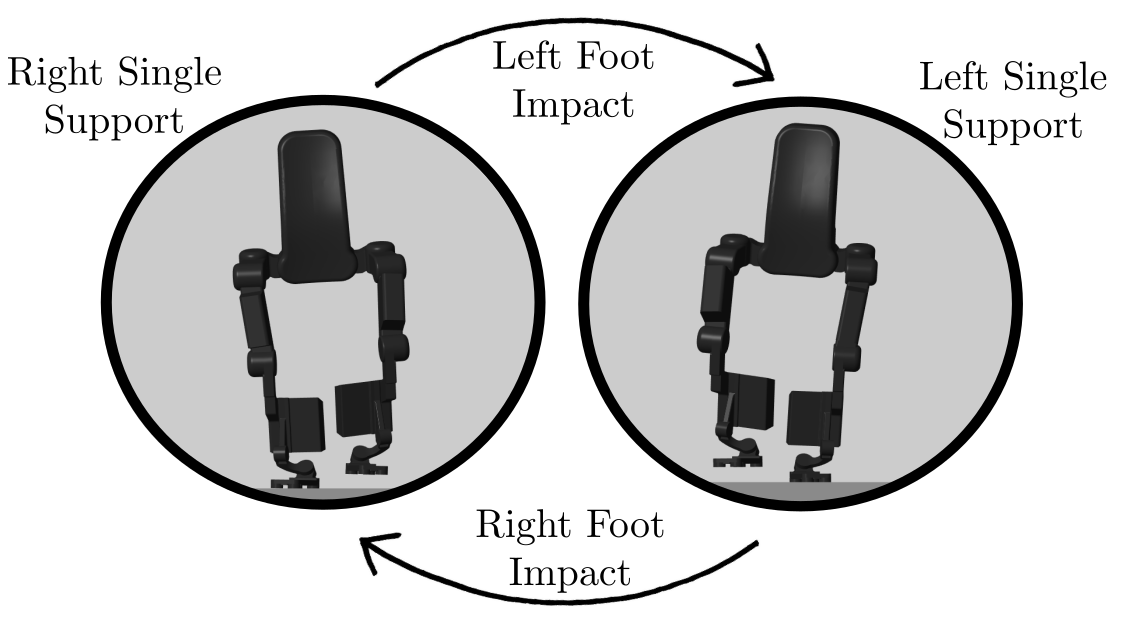}
    \caption{Oriented Graph of Gait}
    \label{fig:OrientedGraph}
\end{figure}

In order to generate an initial gait, the Hybrid Zero Dynamic (HZD) framework with direct collocation was used to find a stable orbit as outlined in \cite{AyongaHZD}. A two domain directed cycle is implemented, with each domain categorized by a single support phase, right and left leg respectively. The two domain representation of the walking gait assumes an instantaneous double-support phase. Transitions between vertices on the graph are triggered by edges when the non-stance foot strikes the ground. The walking behavior of the manikin exoskeleton system is modeled as the hybrid system,
\begin{equation}
    \mathcal{H} \mathcal{C} = (\Gamma, D, U, S, \Delta).
    \label{eq: HZD}
\end{equation}
where $\Gamma$ represents an oriented graph (visually represented in Fig. \ref{fig:OrientedGraph}), $D$ is the domains (right single support and left single support), $U$ represents the admissible inputs determined by the mechanical system, and $\Delta$ is the set of impact maps. 

To simplify the motions, flat footed contact with the ground was imposed. The constraints associated with this contact are taken from \cite{grizzle2014models}. Particularly important to consider are foot slippage and foot roll. Foot slippage is avoided by including a friction cone, explicitly expressed as,
\begin{equation}
    F_z > 0
    \label{eq:fric1}
\end{equation}
\begin{equation}
    F_X^2+F_y^2 \leq \mu^2F_z^2,
    \label{eq:fric2}
\end{equation}

\begin{figure}[t]
    \centering
    \includegraphics[width = 0.8\columnwidth]{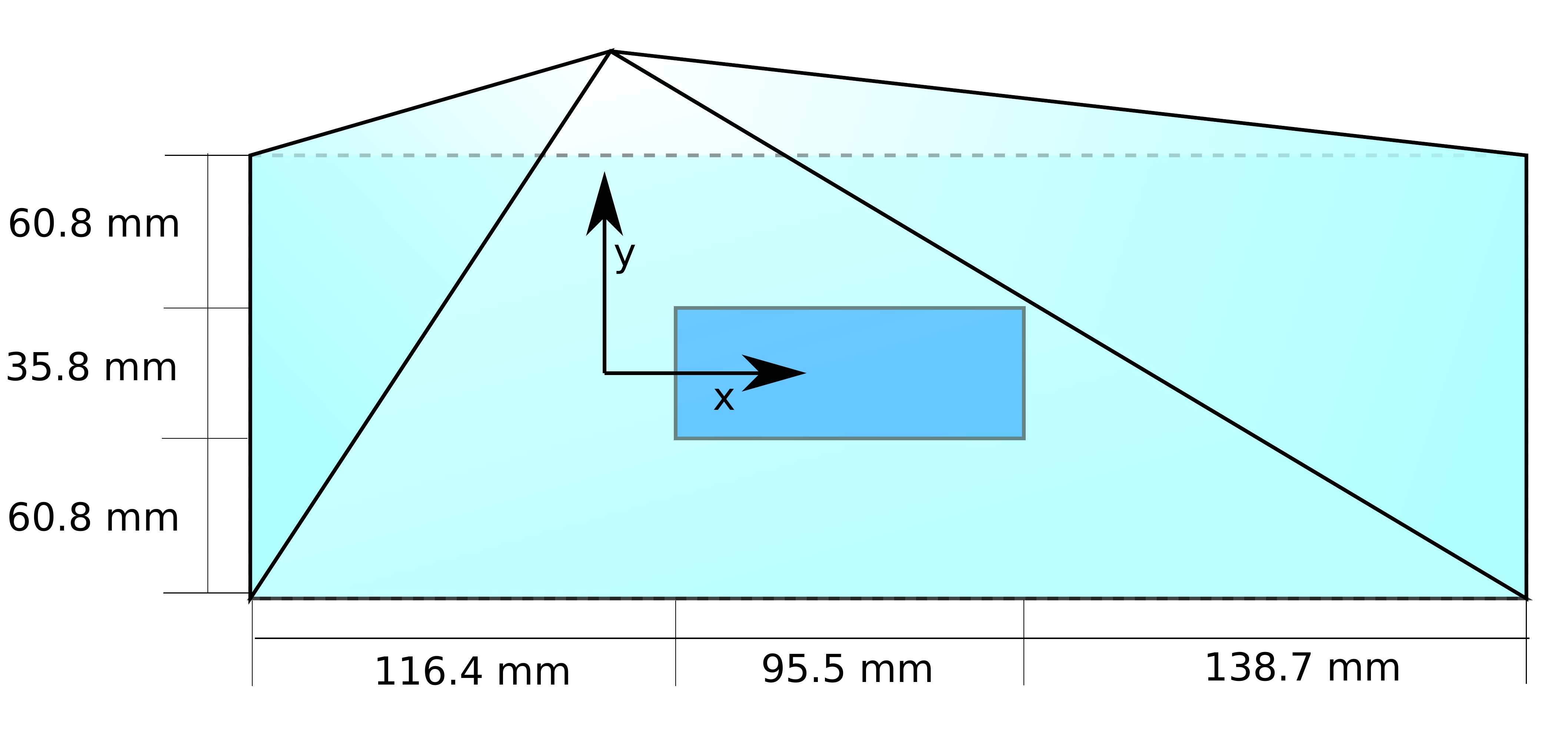}
    \caption{The dark blue area represents the area in which the COP is restricted to on the stance foot. The coordinate frame matches with the world frame in that the $x$direction is to the front of the robot and the y direction is to the left. The left and right direction are symmetric. The coordinates represent the coordinate frame and position of the sole of the foot - directly below actuation}
    \label{fig:ZMP}
\end{figure}

\noindent where the forces are the ground reaction forces on the foot in contact with the ground. Foot roll is contained by limiting the $x$,$y$ and $z$ moments of the same foot. This setup is further discussed in \cite{restoringLocomotion}. In addition, to further ensure robust walking, center of pressure of the stance foot is restricted to a smaller area within the foot. This area is about 96 by 36 mm, and can be seen to scale in Fig. \ref{fig:ZMP}.

Additional constraints were enforced such as foot strike and lift-off velocity, to make convergence easier and the gait healthier for the device. Previous work has designed the gait such that the non-stance foot must go at least 10 cm above the ground in order to avoid early triggering of the impact condition and problems due to terrain irregularities  \cite{feedbackControlofanExoskeleton}. However, it was found that a 6.5 cm margin was sufficient and improved stability. This shorter non-stance foot height enables a smoother, more robust gait. The stable gait, categorized by these constraints as well as the mechanical constraints associated with the robot, is found using the Fast Robot Simulation and Optimization Toolkit (FROST) in MATLAB \cite{FROST}.            
\section{Proposed Stabilization Method}
\label{sec:ProposedMethod}
The control method used most commonly for experimental realization of a stable walking gait is tracking at the joint level, \cite{ames2014human}, \cite{sreenath2011compliant}. However, this method requires a significant amount of time to be spent finding a gait that is both mathematically provably stable and also experimentally successful. The previous experimental results on ATALANTE \cite{restoringLocomotion} also used tracking at the joint level. Although this method achieved a stable walking gait (https://youtu.be/V30HsyUD4fs) it was not robust, and small disturbances quickly caused the system to fall. From these past experiences, it has become clear that a significant disturbance, that frequently causes a loss of balance of the system, occurs when the swing foot impacts the ground at an angle. Thus, the foot hits the ground either with the toe or the heel, rather than directly on the sole of the foot.

\subsection{Analysis of the problem}
\label{subsec:ProblemAnalysis}
\begin{figure}[t]
    \centering
    \includegraphics[width=0.8\linewidth]{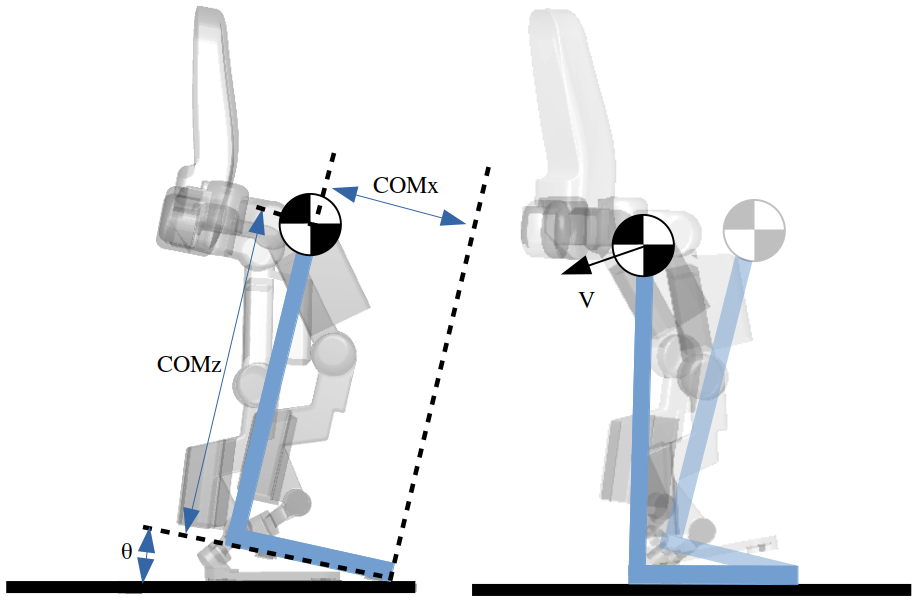}
    \caption{Effect of Impact Angle on Velocity}
    \label{fig:velocitySchematic}
\end{figure}

\begin{figure}[t]
    \centering
    \includegraphics[width = \linewidth]{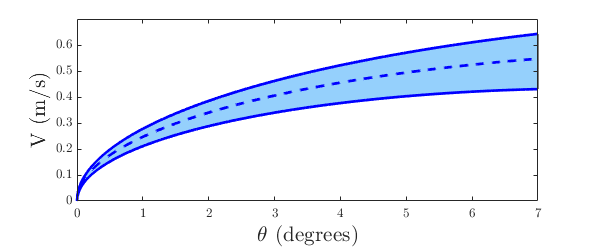}
    \caption{Effect of Impact Angle on Velocity}
    \label{fig:VelocityGraph}
\end{figure}

 As illustrated in Fig. \ref{fig:velocitySchematic}, ``early striking'' of the foot incurs a transfer of potential energy into kinetic energy mostly into an excess (or deficiency) in horizontal velocity of the center of mass. The relationship between error in impact angle and the resulting velocity can be derived using conservation of energy and is as follows:
\begin{align*}
    \begin{bmatrix} \Delta x \\ \Delta z \end{bmatrix} = \begin{bmatrix} \cos(\theta) & -\sin(\theta) \\ \sin(\theta) & \cos(\theta) \end{bmatrix} \begin{bmatrix} \text{COMx} \\ \text{COMz} \end{bmatrix} - \begin{bmatrix} \text{COMx} \\ \text{COMz} \end{bmatrix}
\end{align*}
\begin{align}
    V = \sqrt{2g(\Delta x + \Delta z)}
\end{align}
where $g$ is gravity, COMx is the horizontal distance from the center of mass to the point of impact, and COMz is the vertical distance from the point of impact to the center of mass. Throughout the nominal gait the center of mass is constrained to be within a small area over the center of the foot (Fig. \ref{fig:ZMP}). Thus, we can estimate COMx to be the distance from the toe to within that area. COMz is the vertical distance from the ground to the center of mass when the device is flat on it's feet. The schematic of system at impact can be seen in Fig. \ref{fig:velocitySchematic}.
Fig. \ref{fig:VelocityGraph} illustrates the resulting velocity as a function of impact angle.
It can be seen that even small deviations in the impact angle result in a large deviation from the forward velocity of the system, destabilizing the system. This practically requires very precise tracking of the gait and little disturbance on the system. 

\subsection{Ankle Control} 
\begin{figure}[t]
    \centering
    \includegraphics[width=0.625\linewidth]{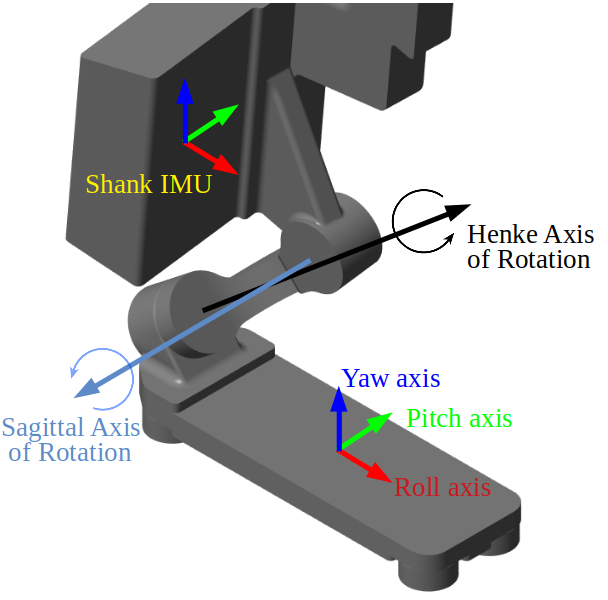}
    \caption{Ankle Coordinate Frames}
    \label{fig:ankleframes}
\end{figure}

The practical solution to preventing the phenomenon discussed in Sec. \ref{subsec:ProblemAnalysis}, and thus obtaining more robust stable walking, is to directly control the orientation of the foot through the Sagittal and Henke ankle joints. The roll, pitch, and yaw axes of the foot as well as the axes of rotation of each ankle joint can be seen in Fig. \ref{fig:ankleframes}.

The main objective of the proposed ankle controller is to ensure that the stance foot remains in contact with the ground and the swing foot is horizontal throughout the step. This will minimize velocity disturbances with respect to the nominal gait. This overall control objective can be broken down into three separate control directives. All three of these control blocks were implemented separately on the exoskeleton during a static one legged stance to illustrate their individual contributions to ankle stability.

\begin{figure}[t]
    \centering 
    \includegraphics[width = \linewidth]{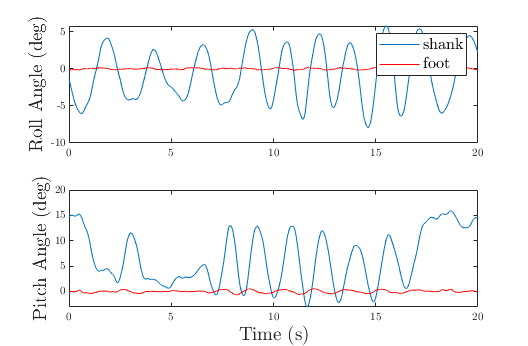}
    \caption{Swing foot roll and pitch compared to the pelvis IMU}
    \label{fig:FlatFeet}
\end{figure}

\noindent \textbf{Keeping the swing foot horizontal.}
The first technique implemented controls the joints of the swing leg ankle to keep the foot horizontal throughout the swing. This was accomplished using traditional inverse kinematics approach. The forward kinematics providing the mapping from joint angles to foot orientation can be express as a product of rotation matrices and lead to the following nonlinear system of equations:
\begin{align}
    \begin{bmatrix} \theta_{roll} \\ \theta_{pitch} \end{bmatrix} = FK(\text{IMU}_{\text{Shank}}, q_{\mathrm{sa}}, q_{\mathrm{ha}})
    \label{eq: invKin}
\end{align}
\noindent where $\theta_{roll}$, and $\theta_{pitch}$ are the roll and pitch angles of the foot with respect to the world frame and $q_{\mathrm{sa}}$ and $q_{\mathrm{ha}}$ are the joint angles of the sagittal and henke ankle joints of the swing foot. Thus $q_{\mathrm{sa}}$ and $q_{\mathrm{ha}}$ are defined as $q_5$ and $q_6$ when in right stance and $q_{11}$ and $q_{12}$ when in left stance. A Newton–Raphson algorithm is used to numerically solve for \eqref{eq: invKin}. This control method was tested experimentally on hardware by manually disturbing the position of the leg and observing the orientation of the foot. This was done when the exoskeleton was hanging from a winch-hoist for ease of moving the leg. The swing foot controller results can be seen in Fig. \ref{fig:FlatFeet}. The disturbance to the system can be seen in the change of roll and pitch angle of the shank, both of which varied by more than 10 degrees. However, as expected, the roll and pitch angle of the foot with respect to the world stayed around zero despite these disturbances. When this controller is applied to the dynamically walking system, the swing foot stays parallel to the ground, preparing it for more ideal and stable impact.

\begin{figure}[t]
    \centering
    \subfloat[][Test on Hardware]{\includegraphics[width=.3\columnwidth]{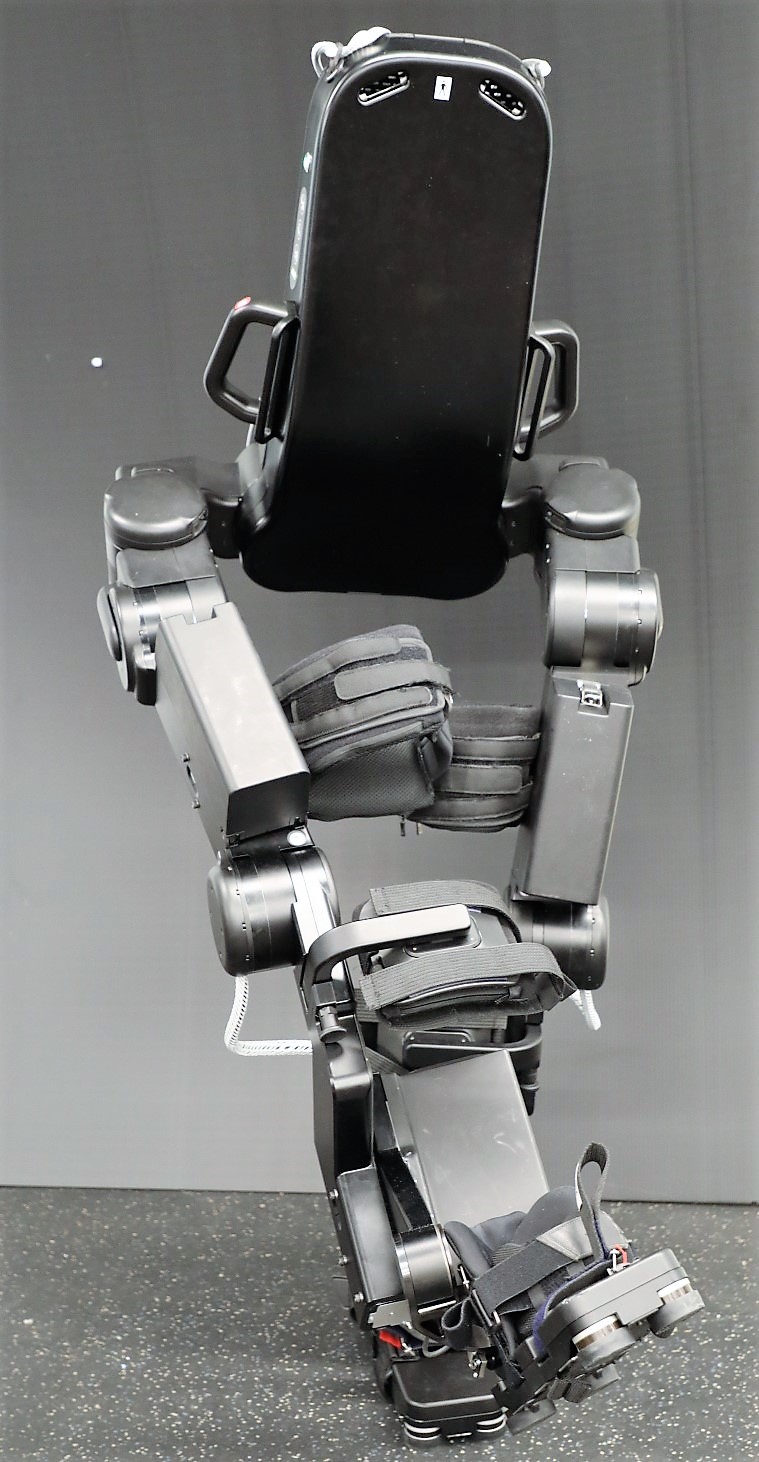}}
    \subfloat[][Simulated Configuration]{\includegraphics[width=.7\columnwidth]{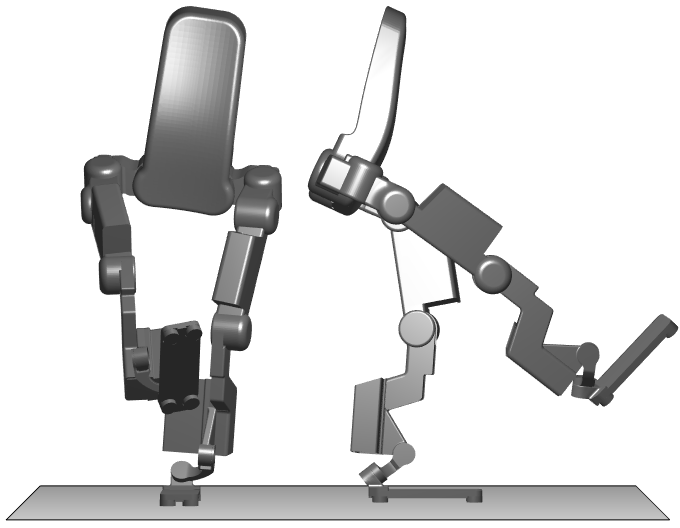}}\\
    \caption{Static Balance Test}
    \label{fig:staticBalance}
\end{figure}

\begin{figure}[t]
    \centering
    \subfloat[]{\includegraphics[width= \columnwidth]{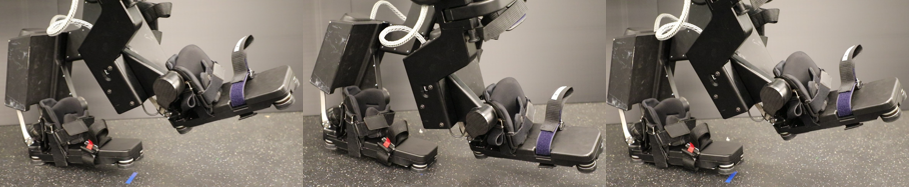}\label{fig:flatfootWithout}}
    \\
    \subfloat[]{\includegraphics[width=\columnwidth]{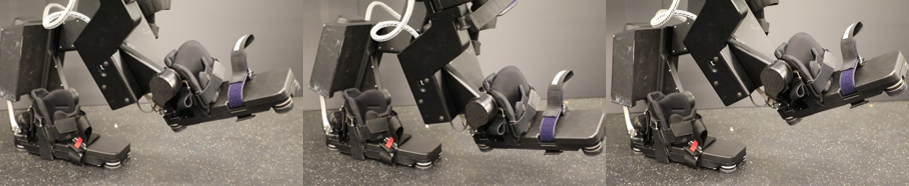}\label{fig:flatfootWith}}
    \caption{(a) Without flat stance foot controller (b) With flat stance foot controller}
\end{figure}

\noindent \textbf{Keeping the stance foot in rigid contact with the ground.}
The second technique implemented ensures that the stance foot maintains rigid contact with the ground. Rigid contact is defined as the condition when the center of pressure (COP) of the foot is located inside the foot. To enforce that, the sagittal and henke ankle joints are saturated based on the measured contact force thanks to sensors located at each corner of the foot. Despite the complex geometry of the ankle (cf. Fig. \ref{fig:ankleframes}), it is possible to saturate its joints in a conservative way by imposing that:
\begin{align}
    q_{sa} &\in Fz[-x_h, x_t] - Fy[0,0] - Fx[z_a, z_a] \\ 
    q_{ha} &\in Fz[-y_i, y_c] - Fy[z_a, z_a] - Fx[0, 0]
    \label{eq: CoPinFoot}
\end{align}
\noindent where $q_{sa}$ and $q_{ha}$ refer to the stance foot, $Fx$, $Fy$ and $Fz$ are the measured reaction forces at center the foot, $x_h$, $x_t$, $y_i$, $y_e$ and $z_a$ are geometrical parameters of the foot and $\alpha$ is a coefficient smaller than one and determined experimentally. This controller was tested experimentally though static balance testing. Static balance refers to achieving stable balance of the exoskeleton with one foot on the ground, and one foot in the air. This balanced position was obtained through solving an optimization program that finds a configuration that constrained the right foot to be flat on the ground, the left foot to be above the ground, and the center of mass to project inside the stance foot. Using these constraints, the optimization problem solved for the position that minimized the torque required to hold this position. The optimal configuration can be seen in Fig. \ref{fig:staticBalance}. 
Disturbances to the one foot balanced configuration without the implemented technique can be seen in Fig. \ref{fig:flatfootWithout}. It can be seen that the stance foot lifted off of the ground due to forces exerted on the pelvis by a human operator. The same disturbance testing was then repeated with the CoP filter activated as shown in Fig. \ref{fig:flatfootWith}. The results showed that the stance foot did not lift off of the ground.

\noindent \textbf{Closing the loop on pelvis orientation.}
It is more representative of disturbances to track the pelvis IMU instead of the ankle joint angles. The method proposed requires the pelvis IMU to be in the correct position, rather than the ankle angles. This enables the walking gait to be robust to deviations in the ground terrain. This counteracts the phenomenon discussed in Sec. \ref{subsec:ProblemAnalysis}, which can also occur if the ground is not perfectly level. 

\begin{figure}[t]
    \centering
    \subfloat{\includegraphics[width= 0.4\columnwidth]{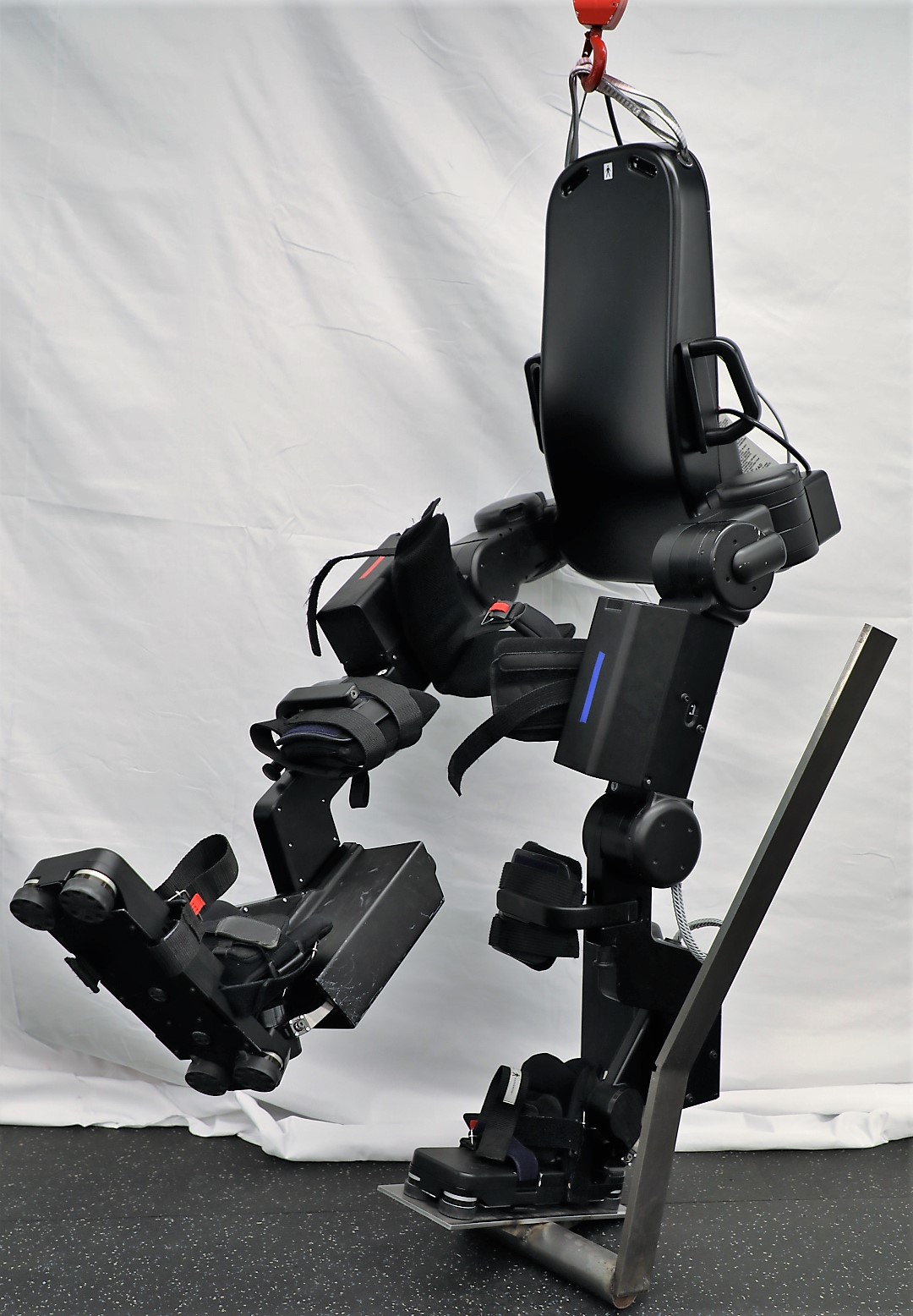}}
    \\
    \subfloat{\includegraphics[width=\columnwidth]{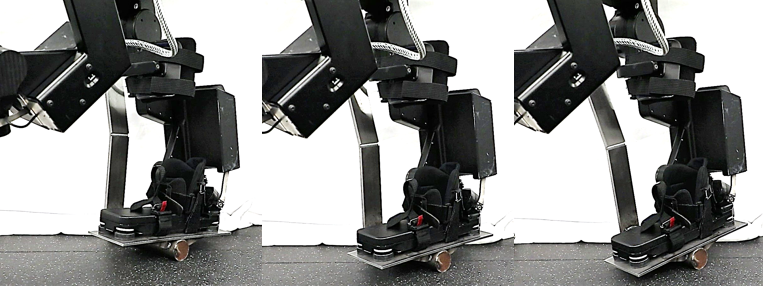}}
    \caption{Exoskeleton balanced on angled platform}
    \label{fig:ExoOnRamp}
\end{figure}

\begin{figure}[t]
    \centering
    \includegraphics[width = \columnwidth]{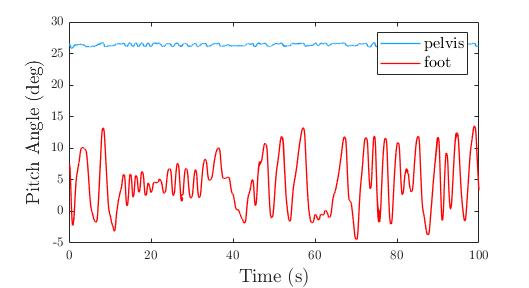}
    \caption{Static Testing of Stance Foot Controller on Angled Platform}
    \label{fig:StaticAngledResults}
\end{figure}

The controller was also experimentally demonstrated using the static balance position showed in Fig. \ref{fig:staticBalance}. The pelvis IMU was tracked as the angle of the ground was disturbed beneath the stance foot. This was accomplished by placing a pivoting platform underneath the stance foot. The experimental setup and results can be seen in Fig. \ref{fig:ExoOnRamp}. It can be seen that the Exoskeleton remains in the static configuration while the pitch of the stance foot is being disturbed. Fig. \ref{fig:StaticAngledResults} further illustrates this by comparing the pitch of the pelvis to the pitch of the foot throughout the testing. It can be seen that the pitch of the pelvis remains fairly constant while the pitch of the foot is changing drastically. These results are significant because they imply that the exoskeleton can accommodate not perfectly even ground.

\section{Stabilization around a Trajectory}
\label{sec:stableTrajectory}

\begin{figure}[t]
    \centering
    \includegraphics[width = \linewidth]{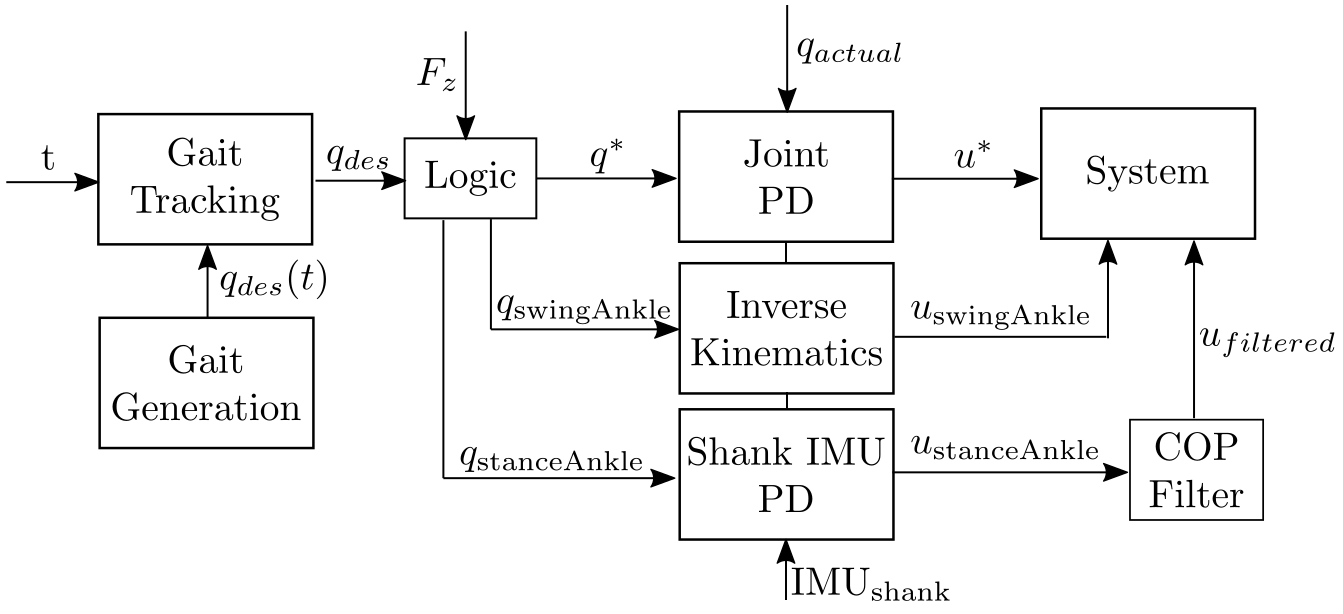}
    \caption{Control Hierarchy}
    \label{fig:hierarchy}
\end{figure}

The proposed components were combined into a single control schematic as shown in Fig. \ref{fig:hierarchy}. 
The generated trajectory, as discussed in Sec. \ref{sec:GaitGeneration}, was commanded as a function of time and tracked using a PD controller. The transition between swing and stance was monitored using a logic condition that triggered an impact when the sum of the z-axis force sensor data crosses a predetermined threshold. A smooth transition between the swing and stance control is applied to avoid abrupt movements that could destabilize the system. The Joint PD controller applies PD control to all non-ankle joints. The ankle controller for the swing foot uses the inverse kinematics method discussed in Sec. \ref{sec:ProposedMethod} to determine the ankle joint angles required to ensure a horizontal foot of the swing leg. The ankle controller for the stance leg calculates the desired pelvis IMU position and tracks it. That control effort is then sent through the COP filter to ensure the commanded torques keep the stance foot rigidly on the ground and the COP in the center of that foot.

\begin{figure}[t]
    \centering
    \includegraphics[width = \columnwidth]{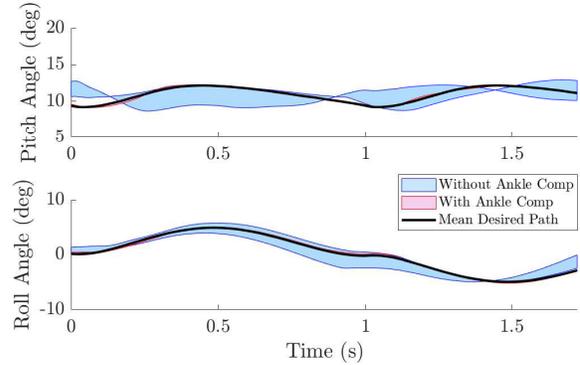}
    \caption{Simulation Results: 4 steps without ankle compensation, 26 steps with ankle compensation}
    \label{fig:simResults}
\end{figure}

\noindent \textbf{Simulation Results} The ankle controller was first tested in simulation. The simulation results can be seen in Fig. \ref{fig:simResults}. The simulated trajectory with no ankle compensation fell after only 4 steps. However after adding the ankle controller, the simulated exoskeleton was able to walk for more than 26 steps with near perfect tracking of the pelvis pitch and roll as shown in Fig. \ref{fig:simResults}.

\begin{figure}[t]
    \centering
    \subfloat{\includegraphics[width = .9\columnwidth]{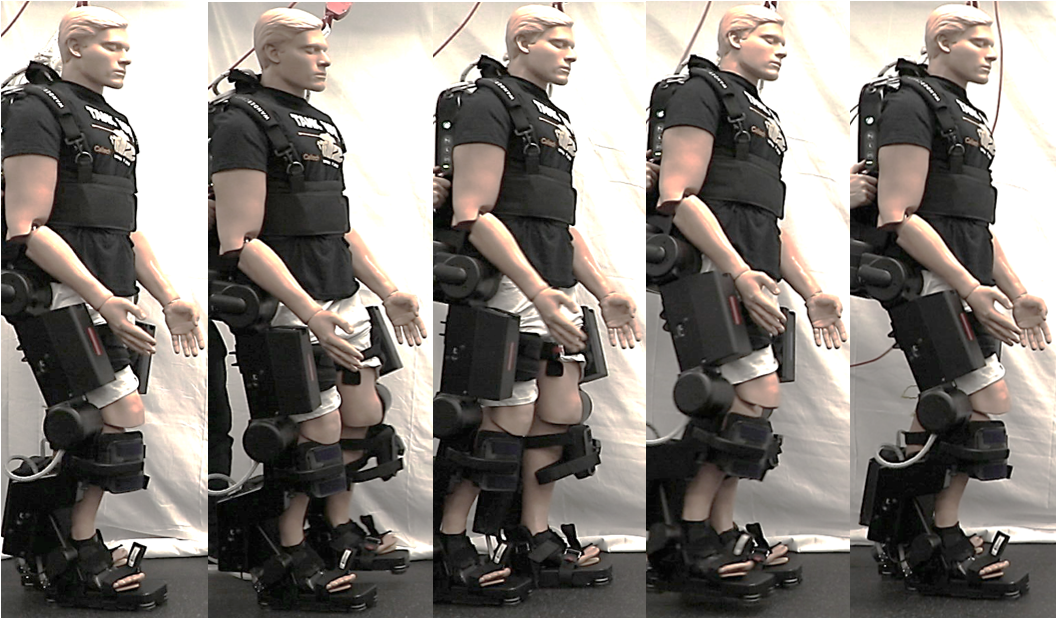}} \\
    \subfloat{\includegraphics[width = .9\columnwidth]{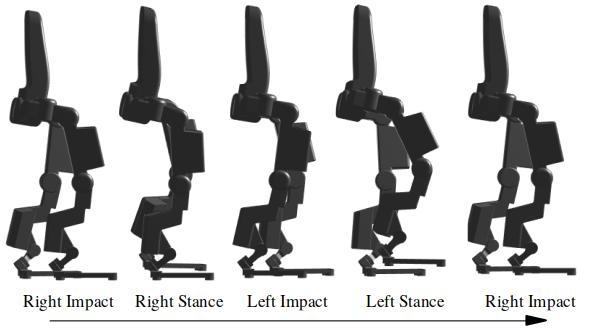}}
    \caption{Nominal Walking Gait}
    \label{fig:nominalTiles}
\end{figure}

\begin{figure}[t]
    \centering
    \includegraphics[width = \columnwidth]{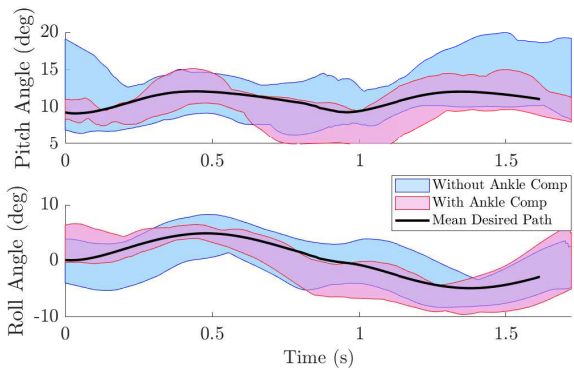}
    \caption{Hardware Results: Both control schemes were run 5 times for 6-8 steps per foot each time}
    \label{fig:experiment}
\end{figure}

\noindent \textbf{Hardware Results} After the simulation results were demonstrated, the ankle controller was implemented on hardware. The experimental realization of the nominal walking gait can be seen in Fig. \ref{fig:nominalTiles}.
Fig. \ref{fig:experiment} demonstrates the increased stability in pelvis orientation in the sagittal plane when the ankle controller is activated. It presents the desired pitch and roll of the pelvis for a standard step - one left step and one right step - as well as the actual pitch and roll of the torso for walking with and without active ankle compensation. The exoskeleton was not able to walk without slight corrections from the operator in the frontal plane. This indicates that despite promising simulation results, real world constraints such as hardware flexibility, approximate modeling, and contact unevenness, this approach does not succeed in stabilizing the system in the frontal plane.

\section{Conclusion}
\label{sec:Conclusion}
In conclusion, this method of active ankle compensation has shown itself to be a valuable tool in sagittal plane stabilization of bipedal robots. The promising individual tests \cite{youtubeclip}, such as achieving static balance on a moving platform, and simulation results indicate that this method has the potential to improve imperfect periodic walking gaits and enhance walking abilities. In practice, these methods did improve certain aspects of walking, such as pelvic pitch tracking (stabilization in the sagittal plane) and foot clearance. However, the stabilization in the frontal plane was unable to be achieved due to physical constraints. Future work could increase robustness to the frontal plane by closing the loop on the pelvis roll through active control of the hips. Utilizing active control of both the ankle and the hip joints suggests the possibility for stabilization on uneven terrain with disturbances in both planes.            
\section*{Acknowledgment}
The authors would like to thank the entire Wandercraft team which designed and constructed ATALANTE, and were valuable resources throughout the process. This material is based upon work supported by the NSF Graduate Research Fellowship No. DGE‐1745301 and NSF NRI Award No. 1724464.        

\bibliographystyle{IEEEtran}
\bibliography{main}

\end{document}